\pdfoutput=1
\documentclass[11pt]{article}
\usepackage{authblk}

\usepackage{acl}

\usepackage{listings}
\usepackage{graphicx}
\usepackage{booktabs}
\usepackage{hhline}
\usepackage{enumitem}
\usepackage{multirow}
\usepackage{hyperref}

\usepackage{times}
\usepackage{latexsym}

\usepackage[T1]{fontenc}
\usepackage[utf8]{inputenc}

\usepackage{microtype}

\title{Investigating the detection of Tortured Phrases in Scientific Literature}

\author[1]{Puthineath Lay}  \author[1]{Martin Lentschat} \author[1]{Cyril Labbé}
  \affil[1]{ Univ. Grenoble Alpes, CNRS, Grenoble INP, LIG, 38000 Grenoble, France}
    \affil[ ]{\textit{puthineath.lay@cadt.edu.kh}, \textit{martin.lentschat@univ-grenoble-alpes.fr}}

\begin{document}
\maketitle
\begin{abstract}
With the help of online tools, unscrupulous authors can today generate a pseudo-scientific article and attempt to publish it.
Some of these tools work by replacing or paraphrasing existing texts to produce new content, but they have a tendency to generate nonsensical expressions.
A recent study introduced the concept of “tortured phrase", an unexpected odd phrase that appears instead of the fixed expression.
E.g. \textit{counterfeit consciousness} instead of \textit{artificial intelligence}. %(i.e. the “expected phrase'').
The present study aims at investigating how tortured phrases, that are not yet listed, can be detected automatically.
We conducted several experiments, including non-neural binary classification, neural binary classification and cosine similarity comparison of the phrase tokens, yielding noticeable results.

\end{abstract}

\section{Introduction}

Scientific texts generated by computer programs can be meaningless, and fake generated papers are served and sold by various publishers with the estimation of 4.29 documents every one million reports  \cite{in_press_cyril}.
But generated texts are also meaningful: with the inputs of a thousand articles, new books are now produced (e.g. \citealp{machine_book}).
Despite the ability of text-generators to produce counterfeit publications, meaningless generated papers can be easily spotted by both machines and humans \cite{torturedphrases}. 
Texts produced by neural language models are more difficult to spot \cite{hutson2021robo}.
These neural language models can produce paraphrased texts that are closer to human-written texts \cite{brown2020language}, and therefore machine-paraphrased texts are harder to differentiate from the human-written texts.

% Optimists could say that those paraphrasing tools are helpful for writers to save time, avoid plagiarism, and learn new ways of writing (\citealp{perez2018technology}; \citealp{fitria2021quillbot}).
Online tools such as \href{https://spinbot.com/}{Spinbot}, and \href{https://www.spinnerchief.com/}{Spinner Chief} are used to paraphrase texts.
However the capacity of a paraphrasing software to assist a writer can be harmful to the scientific literature.
\citet{torturedphrases} screened recent publications (e.g. in the journal \textit{Microprocessors and Microsystems}) and discovered over 500 meaning less phrases in those scientific papers.
They called it "tortured phrases", unexpected odd phrases replacing the lexicalised expression, such as \textit{counterfeit consciousness} instead of \textit{artificial intelligence} (i.e., the expected phrase).
The \href{https://www.irit.fr/~Guillaume.Cabanac/problematic-paper-screener}{database of tortured phrases}, and articles that contain them, have since been expanded to over $9000$ publications in different domains such as Computer Sciences, Biology or Medicine.

In this paper, we investigate strategies to automatically detect new (i.e. unlisted) tortured phrases.
Focusing solely on tortured phrases detection, and not paraphrased text in general,  we will use recent machine learning techniques and state-of-the-art language models.
Our methods were trained on a corpus composed of $141$ known tortured phrases, taking their sentences as contexts, and aims at detecting never-seen-before tortured phrases. 
All code and corpus used are \href{https://gricad-gitlab.univ-grenoble-alpes.fr/nanobubbles/tordetect}{available online}.

% Section \ref{sec:relwork} discuss the related work.
% We introduce a new dataset in Section \ref{extracted_data} and conducted investigation to classify and characterize tortured phrases (cf. Section \ref{sec:exp-res}). 

\section{Related Works}
\label{sec:relwork}

Up to now, no dataset has been built for the automatic detection of tortured phrases.
In \citet{torturedphrases}, authors and contributors collected a set of tortured phrases and their expected phrases that we will use as dataset.
\citet{para_plagiarism} used Spinbot and Spinnerchief to paraphrase original data from several sources such as an arXiv test sets, graduation theses, and Wikipedia articles.
Their study aims at detecting whether a paragraph is machine-paraphrased or not.
The authors tested classic machine learning approaches and neural language models based on the Transformer architecture \cite{vaswani2017attention}, such as BERT \cite{devlin2018bert}, RoBERTa \cite{liu2019roberta}, ALBERT \cite{lan2019albert}, Longformer \cite{beltagy2020longformer}, and others.
They showed that such approaches can complement text-matching software, such as PlagScan and Turnitin, which often fail to notice machine-paraphrased plagiarism.

Because paraphrasing tools like Spinbot and Spinnerchief can generate tortured phrases, the dataset created by \citet{para_plagiarism} surely contains such phrases.
But the task we aim at, i.e. detecting new tortured phrases, is more specific than detecting paraphrased text.
Thus, we investigated three supervised machine learning classifiers: Random Forest, Perceptron, and Transfomer-based model.
Random Forest classifier \cite{breiman2001random} is an ensemble learning method that builds decision trees and classifies each data according to the most selected class.
Perceptron \cite{rosenblatt1958perceptron} is a linear classifier used to classify vectors of numbers.
Term Frequency-Inverse Document Frequency (TF-IDF), GloVe \cite{pennington2014glove}, and BERT \cite{devlin2018bert} were used for word vector representation.
These models were chosen for their state-of-the-art performances and to compare how a model with fixed vectors (i.e. GloVe) compares with a model using dynamic ones (i.e. BERT).

\section{Building datasets of tortured phrases}

This experiment uses two data sources: the tortured phrases \cite{torturedphrases} database and the contexts containing tortured phrases from \citet{para_plagiarism}.
We automatically extracted the context of known tortured phrases from the corpus of \citet{para_plagiarism} to build a training set.

% \paragraph{Tortured phrases}\label{para:human_eval_tortured}

Tortured phrases identified by \citet{torturedphrases} consists of $558$ tortured phrases (e.g. Table \ref{tab:examples}).
These phrases were annotated by the authors and other contributors in several media (e.g. PubPeer, Twitter) in 2021-2022.
After pre-processing, we retained $545$ tortured phrases.

\begin{table}[h]
\begin{tabular}{l|l}
            \toprule
            \textbf{Tortured phrases} & \textbf{Expected phrases} \\ \midrule
            innocent Bayes & naive Bayes\\
            ghostly grouping & spectral clustering\\
            Unused Britain & New England\\
            Joined together states& United States \\
            immature nations & developing countries\\
            \bottomrule 
        \end{tabular}
        \caption{Example of tortured and expected phrases.}
        \label{tab:examples}
    \end{table}

% \paragraph{Machine-paraphrased texts}\label{para:machine_para}

The dataset of \citet{para_plagiarism} is constituted of $193,646$ paragraphs, paraphrased using Spinbot and Spinnerchief.
$65,433$ original data were retrieved from several sources: arXiv, graduation theses of ESL students at the Mendel University in Brno (Czech Republic) and Wikipedia articles. 
%For our experiment, we use the paraphrased test corpus as our dataset.

% \paragraph{Extracted paragraph}\label{extracted_data}

We extracted the paragraphs containing tortured phrases \cite{torturedphrases} to build a training and evaluation corpus. % referenced in \ref{para:human_eval_tortured}.
This resulted in $1,104$ paragraphs with tortured phrases and $1,668$ paragraphs without tortured phrases (randomly extracted from the non-paraphrased original data). % (i.e. the data before getting paraphrased).

\paragraph{Data augmentation: five-grams extraction}

To increase the training data, we extracted the n-grams (i.e. sequences of n adjacent tokens) of each sentence, with $n=5$, as it is the maximum length of the known tortured phrases.
A five-gram is considered positive if a complete tortured phrase appears in that five-gram.%, otherwise it will be considered negative.
This produced $38,397$ five-grams, $5,024$ positive five-grams (in the '1' class) and $33,373$ negative five-grams (in the '0' class).

\section{Experiment and Result}
\label{sec:exp-res}

% \subsection{Classification tasks}
We investigated binary classifiers to check the difference between paragraphs or five-grams containing tortured phrases in several settings.
The paragraphs and five-grams containing tortured phrases are considered positives, with label '1', while negative paragraphs are labeled as '0'.
Accuracy, precision, recall, and F-measure are used to evaluate the classification performances.

\paragraph{Classifiers: Random Forest and Perceptron} 

In this experiment, five-grams data are used in the classification. 
The five-grams are converted to a numerical representation using Sklearn TF-IDF count vectorizer and split randomly 80\% for training and 20\% for testing.
We used the Scikit Learn library for the \href{https://scikit-learn.org/stable/modules/generated/sklearn.ensemble.RandomForestClassifier.html}{Random Forest} and \href{https://scikit-learn.org/stable/modules/generated/sklearn.linear_model.Perceptron.html}{Perceptron} with the default value of all parameters.

The result in Table \ref{tab:results} shows an accuracy for the Random Forest classifier of $.98$ and the Perceptron of $.94$.
The precision, recall, and F1-score of the Random Forest classifier are high, especially in class $0$, and the results in class $1$ is slightly lower than in class $0$.
The precision, recall, and F1-score of the Perceptron method is slightly lower than that of the Random Forest classifier, but it is still comparable for class $0$.
In class $1$, Perceptron results are significantly lower compared to the results of Random Forest classifier.
We also observe results higher in class $0$ than in class $1$, this might be due to the data imbalance.

After observing the accuracy, precision, recall, and F1-score, we see that the models perform well based on TF-IDF vector representation.
However, it is believable that the models learned to classify five-grams based solely on specific words: since the training and test data were split randomly, a tortured phrase can be present in both sub-sets.

% In the next experiment, we worked on Transformer-based model to check the performance of the tortured phrases in the paragraphs or in the five-grams. 

\paragraph{Transformer-based classifier on paragraphs}
\label{subsec:tran-para}

Here, we seek at detecting paragraphs containing at least one tortured phrase.
The data are split 67\% for training and 33\% for test set.
The architecture of this model is based on the Transformer technique.
Pre-trained transformers from Huggingface, \href{https://huggingface.co/docs/transformers/model_doc/distilbert}{\texttt{distilbert-base-uncased}} model \cite{distilbert}, was chosen for its lightness and speed.
We applied transfer learning by adding one linear layer for classification purposes.
In that linear layer, the number of input features was set to $768$ with an output size of $2$, indicating class $0$ and class $1$.
The model was trained on $10$ epochs.

The results in Table \ref{tab:results} show an accuracy of $.86$.
The $.92$ precision on class $1$ is higher than on class $0$.
For the recall and F1-score, class $0$ gets a better result than class $1$.
Since the amount of paragraph data are small, we suspect that only a few tokens constitute the tortured phrases, and that the rest of the token's paragraph affect the performance.

\paragraph{Transformer-based classifier on five-grams}

The data of class '1' was split 79\% for training and 21\% for testing by filtering the test set with tortured phrases not present in the training set.
In this experiment, two versions of the model were trained: one using the entire dataset and one with a proportion of data balanced in both classes.

The training data are made of $28,995$ five-grams ($25,029$ in class '0' and $3,966$ in class '1').
The test data are made of $9,402$ five-grams ($8,344$ and $1,058$).
Table \ref{tab:results} shows an accuracy of $.88$.
Precision, recall, and F1-score on class '1' are exceptionally low compared to the '0' class.

Regarding the classifier with balanced data, the size of the training set is $7,932$ five-grams ($3,966$ in each class) and the size of the testing set is $2,116$ five-grams ($1,058$ in each class). The accuracy is $.71$, and the precision, recall, and F1-score are around $.70$ in both classes (cf. bold values in Table \ref{tab:results}).
In this experiment, the balance of the classes induces a greater reliability of the results and we believe this approach presents the best applicability. The model focuses on the tortured phrases in five-grams rather than tortured phrases in the whole paragraph, so the model can learn to generalize the five-grams containing tortured phrases or not.%, which produces the best score (if we compare the score of all four metrics) among all classification experiments.}

\begin{table*}[htbp]
\centering
\begin{tabular}{l|l|c|cc|cc|cc}
\hline
\hline
\multicolumn{1}{c|}{Classifiers} & \multicolumn{1}{c|}{Data type}                                             & Accuracy             & \multicolumn{2}{c|}{Precision}               & \multicolumn{2}{c|}{Recall}                  & \multicolumn{2}{c}{F1-score}                 \\ 
\cline{1-1}\cline{4-9}
% \cline{1-9}
\multicolumn{1}{c|}{class}       & \multicolumn{1}{c|}{}                                                      &                      & \multicolumn{1}{c}{0} & 1                    & \multicolumn{1}{c}{0} & 1                    & \multicolumn{1}{c}{0} & 1                   \\ 
\hline
\hline

Random Forest                  & Random five-grams                                                              & .98                 & .99                  & .92                 & .99                  & .91                 & .99                  & .92                   \\

Perceptron                     & Random five-grams                                                              & .94                 & .96                  & .84                 & .98                  & .69                 & .97                  & .75                   \\ 
\cline{1-9}

Transformer                   & Paragraph                                                                  & .86                 & .82                  & .92                 & .94                  & .77                 & .87                  & .84                 \\ 

Transformer                    & \begin{tabular}[c]{@{}l@{}}Random five-grams \end{tabular}   & .88                 & .89                  & .42                 & .99                  & .03                 & .93                  & .06                   \\

Transformer                    & \begin{tabular}[c]{@{}l@{}}Balanced five-grams \end{tabular} & \textbf{.71}        & \textbf{.67}         & \textbf{.75}        & \textbf{.79}         & \textbf{.62}        & \textbf{.73}         & \textbf{.68}          \\
\hline
\hline
 
\end{tabular}
\caption{Classification results. }%{\color{blue} The bold values indicates the best score among all experiments.}}
\label{tab:results}
\end{table*}

\subsection{Cosine similarity comparison}
We studied the cosine similarity between tokens in the tortured phrases compared to the similarity of tokens in the expected phrases. 
\citet{torturedphrases} annotated a dataset of tortured phrases, and their respective expected phrases, from several media such as PubPeer during 2021-2022.

We intuitively expect that the cosine score of the tokens in the phrases could yield noticeable results, useful to differentiate tortured and expected phrases due to the similarity, or non-similarity, of adjacent tokens.
The expected phrases are idioms (i.e. multi-word-expression forming a lexical and semantic unit) and, as such, we hypothesize that the semantic score defined by the cosine of the vectors between their terms should be higher than for the tortured phrases, which words are less likely to be semantically related or frequently associated.
If validated, such observation could help distinguish tortured phrases from legitimate ones.
For this experiment, only the similarity of two-tokens phrases were computed, using two kinds of word embedding models: GloVe and BERT. 

\paragraph{Cosine similarity on phrases using BERT} 
In this study, we used BERT \cite{devlin2018bert} as the word embedding model. We followed the architecture of \citet{mccormick_2019} by summing the last four layers of 12 layers of BERT to get one-word vectors with $768$ values. 

Since the BertTokenizer will separate unknown words into sub-words (e.g. \textit{vitality utilize} becomes \textit{vital}, \textit{\#ity}, and \textit{utilize}), it can be complicated to compute their cosine similarity.
We chose to discard tortured phrases containing words unknown by the model.
We retained  $82$ tortured phrases after the tokenization process. 

\begin{figure}[h]
\centering
\includegraphics[width=.5\textwidth]{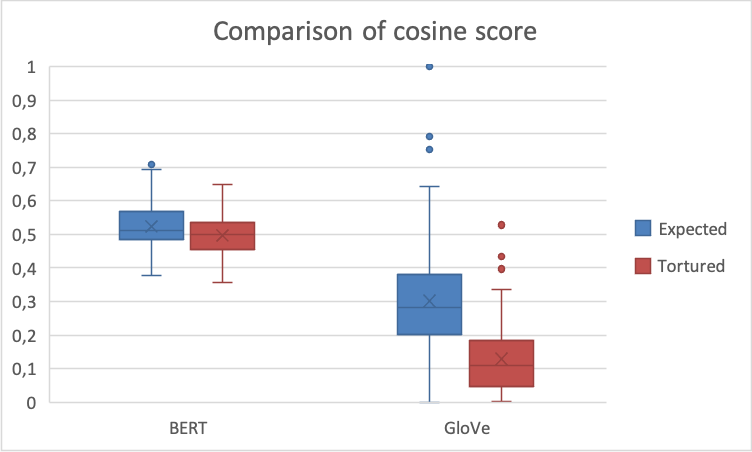}
\caption{Comparison of cosine score of phrases using BERT and GloVe.}
\label{fig:bert-glove-cosine}
\end{figure}

The scores obtained from cosine computation between word pairs in tortured and expected phrases present slight differences, as shown in Figure \ref{fig:bert-glove-cosine}.
The median scores of expected and tortured phrases are $.51$ and $.49$, respectively.
The absence of significant differences can be explained by the nature of the BERT model: a two-word context is probably not sufficient to differentiate tortured and expected phrases using cosine similarity.

\paragraph{Cosine similarity on phrases using GloVe}

For this experiment, we computed the cosine similarity of token pairs in tortured and expected phrases. We used the pre-trained GloVe word embedding \cite{pennington2014glove} (\texttt{glove.6B.200d.txt})%, which refers to the 6B tokens, 400K vocab, uncased and 200d vectors model
which was trained on Wikipedia 2014 and \href{https://catalog.ldc.upenn.edu/LDC2011T07}{Gigaword}.
Unlike BERT, GloVe is a context-free model, meaning that each word in this pre-trained model is assigned to one constant vector.
However, GloVe vocabulary is limited and thus some tokens in the phrases might not appear in this model.
For this issue, we padded the out-of-vocabulary word with $0$.
For the phrases in which both words do not appear in the model, the cosine similarity is $0$.
We discarded the phrases whose scores were lower than or equal to $0$ ($cosine\_score \leq 0$).
As a result, 139 phrases are used for this experiment.

The Figure \ref{fig:bert-glove-cosine} indicates that the cosine similarity scores of tortured phrases tend to be smaller than those of expected phrases when using GloVe. The median score of expected phrases is $.3$ and the median score of tortured phrases is $.12$.

These results indicate that the cosine score between terms could be employed to differentiate tortured phrases from legitimate ones.
Since Bert relies on embeddings, it is overly influenced by the phrase contexts to yields useful results.
With Glove, or another language model with static vectors, one could chose a threshold to classify phrases, e.g. as legitimate, tortured, or requiring human expertise.

% \subsection{Discussion}
% % will give insight on RF and perceptron
% For the classification tasks, the results in accuracy of all Transformer-based classification approaches is around 70\%.
% Transformer-based classifier with the balanced training and test set of five-gram data yields the best result among all classification experiments.
% Regarding the cosine similarity computation, the variance of the phrase pairs using GloVe word embedding is more significant than that of phrase pairs using BERT word embedding.
% BERT takes context into account, while GloVe does not.
% The Figure \ref{fig:bert-glove-cosine} shows the average score of tortured phrases and expected phrases comparison.
% For this experiment, using GloVe tends to help us differentiate the characteristics of tortured and expected phrases due to the small similarity of tokens in tortured phrases. 

% maybe put in appendix

\section{Conclusions}\label{sec:conc}

In this research, we aimed at detecting new tortured phrases.
We studied different classification approaches and examined the characteristics of tortured phrases using cosine similarity.
The result of Perceptron and Random Forest classifier are high, but we intuitively suspect they are not reliable due to the word representation using TF-IDF vectorization.
The Transformer-based classifier model with paragraph data provided the best result among Transformer models.
However, we suspect that the model learned to classify paragraphs based only on a few tokens% data contain only a few tokens as the tortured phrases, so the rest of the tokens that are not tortured phrases in the paragraph can affect the performance of how the model detects the positive or negative paragraphs.
, and classifying paragraphs is not sufficient to detect the exact tortured phrases.
Thus, the Transformer-based classifier model five-gram data yields the best result with balanced classes (i.e. results above $.70$ for all metrics).
%The four metrics (i.e. accuracy, recall, precision, F1-score) of this model are above $.70$ for each of the two categories.

We also studied the use of cosine similarity between the phrase tokens to identify new tortured phrase.
This showed that language model with fixed vectors (e.g. Glove) could be used to classify part of the phrases. 
%, e.g. using the median score as threshold (due to the small cosine score of tortured phrases using GloVe).

Future research should include more human evaluation of tortured phrases and a bigger dataset tortured phrases with their context. 
To improve the classification, future work could investigate Support Vector Machine (SVM) and Na\"ive Bayes model (NB): SVM performs better with a small dataset and binary class, while NB can provide probabilities of a prediction.% with knowledge appears in the training data.
% Another perspective would be to work with Seq2Seq language model, to detect the tortured phrases and deduce the expected phrases.
Finally, computing the cosine similarity of the tortured phrase and expected phrase pairs within the whole context to see tortured phrases' performance.

\section*{Acknowledgments}
The \href{https://nanobubbles.hypotheses.org/}{NanoBubbles} project has received Synergy grant funding from the European Research Council (ERC), within the European Union’s Horizon 2020 program, grant agreement no. 951393.

\bibliography{custom}
\bibliographystyle{acl_natbib}

% \appendix

% \section{Example Appendix}
% \label{sec:appendix}

% This is an appendix.

\end{document}